\documentclass[runningheads]{llncs}
\usepackage{graphicx}
\usepackage{graphicx}
\usepackage{amsfonts}
\usepackage{amsmath}
\usepackage[linesnumbered, lined, ruled,vlined]{algorithm2e}
\usepackage{wrapfig}
\usepackage{amsmath}
\usepackage[colorinlistoftodos]{todonotes}
\usepackage{epsfig}
\usepackage{amssymb}
\usepackage{booktabs}
\usepackage[mathscr]{eucal}
\usepackage{xcolor,colortbl}
\usepackage{bigints}

\DeclareMathAlphabet{\mathcal}{OMS}{cmsy}{m}{n}

\def\argmin{\mathop{\rm argmin}}

\newcommand{\inner}[2]{\left\langle #1,#2 \right\rangle}

\newcommand{\real}{\ensuremath{\mathbb{R}}}

\newcommand{\ltwo}{\ensuremath{\mathbb{L}^2}}

\newcommand{\norm}[1]{\left\lVert #1 \right\rVert}

\newcommand{\fn}{\mathcal{F}}
\newcommand{\bundl}{\mathcal{B}} 
\newcommand{\aproj}{\mathcal{A}}
\newcommand{\Orot}{\mathcal{O}}

\usepackage{bbm}

\begin{document}
\title{Alignment of Tractography Streamlines  using Deformation Transfer via Parallel Transport}
\titlerunning{Alignment of Tractography Streamlines using ...}
\author{Andrew Lizarraga\inst{1} \and
David Lee\inst{1,3} \and
Antoni Kubicki\inst{1} \and
Ashish Sahib\inst{1} \and
Elvis Nunez\inst{1,2} \and
Katherine Narr\inst{1,4} \and
Shantanu H. Joshi\inst{1,3}
}

\authorrunning{A. Lizarraga et al.}

\institute{Ahmanson Lovelace Brain Mapping Center, Department of Neurology, UCLA \and
Department of Electrical and Computer Engineering, UCLA \and
Department of Bioengineering, UCLA \and
Department of Psychiatry and Behavioral Sciences, UCLA\\
\email{AndrewLizarraga@mednet.ucla.edu}, \email{dalee@mednet.ucla.edu}, \email{akubicki@ucla.edu}, \email{asahib@ucla.edu}, \email{elvis.nunez@ucla.edu}, \email{knarr@mednet.ucla.edu}, \email{s.joshi@g.ucla.edu}
}
\maketitle              % typeset the header of the 
\begin{abstract}
We present a geometric framework for aligning white matter fiber tracts. By registering fiber tracts between brains, one expects to see overlap of anatomical structures that often provide meaningful comparisons across subjects. However, the geometry of white matter tracts is highly heterogeneous, and finding direct tract-correspondence across multiple individuals remains a challenging problem. We present a novel deformation metric between tracts that allows one to compare tracts while simultaneously obtaining a registration. To accomplish this, fiber tracts are represented by an intrinsic mean along with the deformation fields represented by tangent vectors from the mean. In this setting, one can determine a parallel transport between tracts and then register corresponding tangent vectors. We present the results of bundle alignment on a population of 43 healthy adult subjects.

\end{abstract}
%
%
%

%Zhengwu2019nonparametric
%prasad2014automatic
\section{Introduction}
There have been numerous approaches to overcome the difficulty of inter-subject comparison of white matter fiber tracts. Glozman et al. presented a compact geometric model of white matter fibers composed of a center-line and groups of point clouds along the length of the center-line to drive registration and analysis across the subjects~\cite{glozman2018framework}. While this could potentially provide an efficient and effective representation and registration, it still may not fully account for the rich geometry within the fiber streamlines. Zhang et al. noted that statistical methods tend to disregard the rich geometry of fiber bundles and  proposed to model fibers as a non-parametric Bayesian process which captures the overall geometry and allows for statistical analysis~\cite{Zhengwu2019nonparametric}. However, the results in this model are not necessarily deterministic.
Durrelman et al. proposed an elegant solution by representing the streamlines as currents embedded in the vector fields. This allows estimation of diffeomorphic\index{diffeomorphic} transformations between two sets of fiber bundles thus capturing the geometric variability across the sets of streamlines~\cite{durrleman2011registration}. Garyfallidis et al. introduced a framework for streamline-based registration of two fiber tracts without introducing simplified models, and instead solved for a linear transformation that minimizes the cost function of distances between a pair of streamlines\index{streamlines}~\cite{garyfallidis2015robust}. Prasad et al. proposed a maximum density path representation for capturing  geometric information of bundles and showed the benefit of automatically segmenting bundles into regions of interest by virtue of the density paths respecting geometrically distinguished portions of the bundle~\cite{prasad2014automatic}.
Another approach by O'Donnell et al. has used distributions of fibers for unbiased group registration of tractography\index{tractography} representations~\cite{o2012unbiased}. A recent approach by Zhang et al. presented a rich and detailed set of white matter fiber atlas, which allowed direct registration of whole brain white matter fibers, as well as clustering into major fiber bundles~\cite{zhang2018anatomically}. 

\subsection{Contributions of the Paper}
In this paper, we rely on parameterized representations of fiber streamlines in an appropriate geometric space with the Fisher-Rao metric~\cite{rao1945information}, along with a low dimensional tangent vector representation of the fibers. This low-dimensional representation takes inspiration from the tangent space principal component analysis developed by Joshi et al.~\cite{JOSHI2013547}. We propose using this representation, as it reduces the problem of registration to a problem of parallel transport\index{parallel transport} and optimal rotation\index{optimal rotation} of the low-dimensional tangent vectors\index{low dimensional tangent vectors}.

In this paper the following contributions are made: i) a new representation that encodes the underlying geometry (mean) of the fiber bundle along with low-dimensional representations of individual fibers about the mean, ii) a new metric in the product space of the mean and the tangent space projections, and finally iii) an alignment method for registering the means and low-dimensional tangent vectors. We show that this process of registration corresponds to tract alignment in the original space. We present the results of bundle alignment on $N=43$ healthy controls and show significantly lower discrepancy in matching streamlines compared to standard rigid-alignment. 

\section{Methods}
\subsection{Preliminary Background}

The coordinates of a white matter fiber tract with $N$ fibers can be represented by a set of continuous functions in a Hilbert space, $\fn \equiv \{(f_1, f_2, \ldots, f_N)\mid f_i \in \ltwo([0, 1], \real^3)\}$. We sought for a bundle representation that is general and encodes the shape variability of the fiber bundle around the mean. While there are several choices for the shape representations of $\{f_i\}$ including the well-known current-based representations, in this paper we use parameterized curves. Specifically, we use the square-root velocity function\index{square root velocity function} (SRVF) to represent the geometry of the fibers~\cite{Joshi2007,Joshi2007a,srivastava_etal_PAMI:11}. Each fiber $f_i$ is represented by a SRVF map $f_i \mapsto q_i$, where $q_i = \dot{f_i}/\sqrt{||\dot{f_i} ||}$, and imposing the constraint $\int_{[0,1]} \inner{q(t)}{q(t)} dt = 1$ ensures that our shape representations are invariant to translation and scaling. We denote the space of such shapes as $\cal S$. The tangent space of shapes $T_q({\cal S})$ under this representation is endowed by a Fisher-Rao metric which is the standard $\ltwo$ inner product invariant under domain reparameterizations of $q_i$.

\subsection{Fiber Tract Representation}
We represent the shape of the fiber tract as a pair $\bundl \equiv (\beta_\mu, \aproj)$, where $\beta_\mu$ is the mean shape of $\{q_i \}$, and $\aproj$ is a low-dimensional projection of the tangent vectors from the mean shape $\beta_\mu$ in the direction of the shapes $q_i$. The first component of this representation is computed as the K{\"a}rcher mean~\cite{karcher77}\index{karcher mean},
\begin{equation}
\beta_\mu=\argmin_{q} \frac{1} {N} \sum_{i=1}^{N}  \argmin_{O, \gamma} \norm{ q - \sqrt{\dot{\gamma}} \Orot q_i (\gamma)}^2,
\label{eq:betamu}
\end{equation}
where $\Orot \in SO(3)$, and $\gamma :[0, 1] \to [0, 1]$ is a reparameterization function.

To derive the second component, we consider a set of tangent vectors $\{v_i \mid v_i \in T_{\beta_\mu}({\cal S}), i = 1, \ldots, N$\}, such that the exponential maps $\mbox{exp}_{\beta_\mu}(v_i) = q_i$ recover the set of $\{ q_i\}$. This map is defined as $\mbox{exp}_{\beta_\mu}(v_i) = \beta_\mu \cos \alpha + f \sin \alpha$, where $\cos \ \alpha = \langle \beta_\mu, q_i\rangle$ and the initial tangent vector is given by $f=q_i - \langle \beta_\mu, q_2\rangle$. While there are several choices of basis for $T_{\beta_\mu}({\cal S})$, in this work we adopt a Fourier basis. We denote the basis elements as $\{ g_k \}, k=1,\ldots, N$ and project them on $T_{\beta_\mu}({\cal S})$ as  $G \equiv \{  \tilde{g}_k \}$, where $\tilde{g}_k = g_k - \langle g_k, \beta_\mu \rangle \beta_\mu$,  $k=1,\ldots, N$. Then the second component of the bundle representation $\aproj$ is given by
\begin{equation}
    \aproj = \langle v_i, g_k\rangle, i = 1, \ldots, N, k = 1, \ldots, N
    \label{eq:aproj}
\end{equation}
Finally, the fiber tract is represented by the pair of functions,
\begin{equation}
    \bundl \equiv (\beta_\mu, \aproj),~ \beta_\mu \in {\cal S},~ \aproj \in \ltwo([0, 1], \real^{N \times N}).
    \label{eq:bundle_rep}
\end{equation}

Note that the term $\aproj$ can be interpreted as a coefficient matrix, with the $i$-th row as the set of Fourier coefficients of the $i$-th tangent vector, and we treat these coefficients as shape features. Additionally, the rows of $\aproj$ are in one-to-one correspondence with the tangent vectors. As a consequence, one can change the direction and scale of the tangent vectors by adjusting these coefficients. Since we treat an entire tract as its mean paired with its low-dimensional tangent vectors, the most natural deformation to perform on the $\aproj$ while respecting the mean are rotational actions in $SO(N)$.

\subsection{Approach to the Registration Problem}
We provide a rationale to our approach in tract registration: Note that computing the Fisher-Rao metric between K\"archer means, $\inf_{\gamma}||\beta_{\mu}^i - \sqrt{\dot{\gamma}}\beta_{\mu}^j(\gamma)||$, gives rise to a geodesic between the respective means. This warp is used to register the means between two tracts, by traversing the geodesic from one mean to the other.

However, if one pursues further, we can parallel transport the tangent vectors in $T_{\beta_\mu^i}({\cal S})$ to the tangent space $T_{\beta_\mu^j}({\cal S})$ along the mean geodesic, which is described in detail in Alg.\ref{algo:parallel_tx} . After transport, successful alignment of the two sets of tangent vectors correspond to a registration between the two tractography streamlines. Given the tangent space is high dimensional, we extract the low dimensional representation of the two sets of tangent vectors as described in Eqn. \ref{eq:aproj}. An approximate registration then boils down to finding an appropriate rotation between the low-dimensional tangent spaces. The optimal rotation between two low-dimensional features $\aproj^1$ and $\aproj^2$, can be found by taking a singular value decomposition of the product $\aproj^1 {\aproj^2}^T = R \Sigma Q^T$, where $R, Q \in SO(N)$. Then the optimal rotation matrix $\mathcal{O}$ is given by $\mathcal{O} = R Q^T$~\cite{goryn1995estimation}.

\subsection{Fiber Bundle Distance}

 \begin{figure}[tbh]

\centering
     \includegraphics[width=0.67\textwidth]{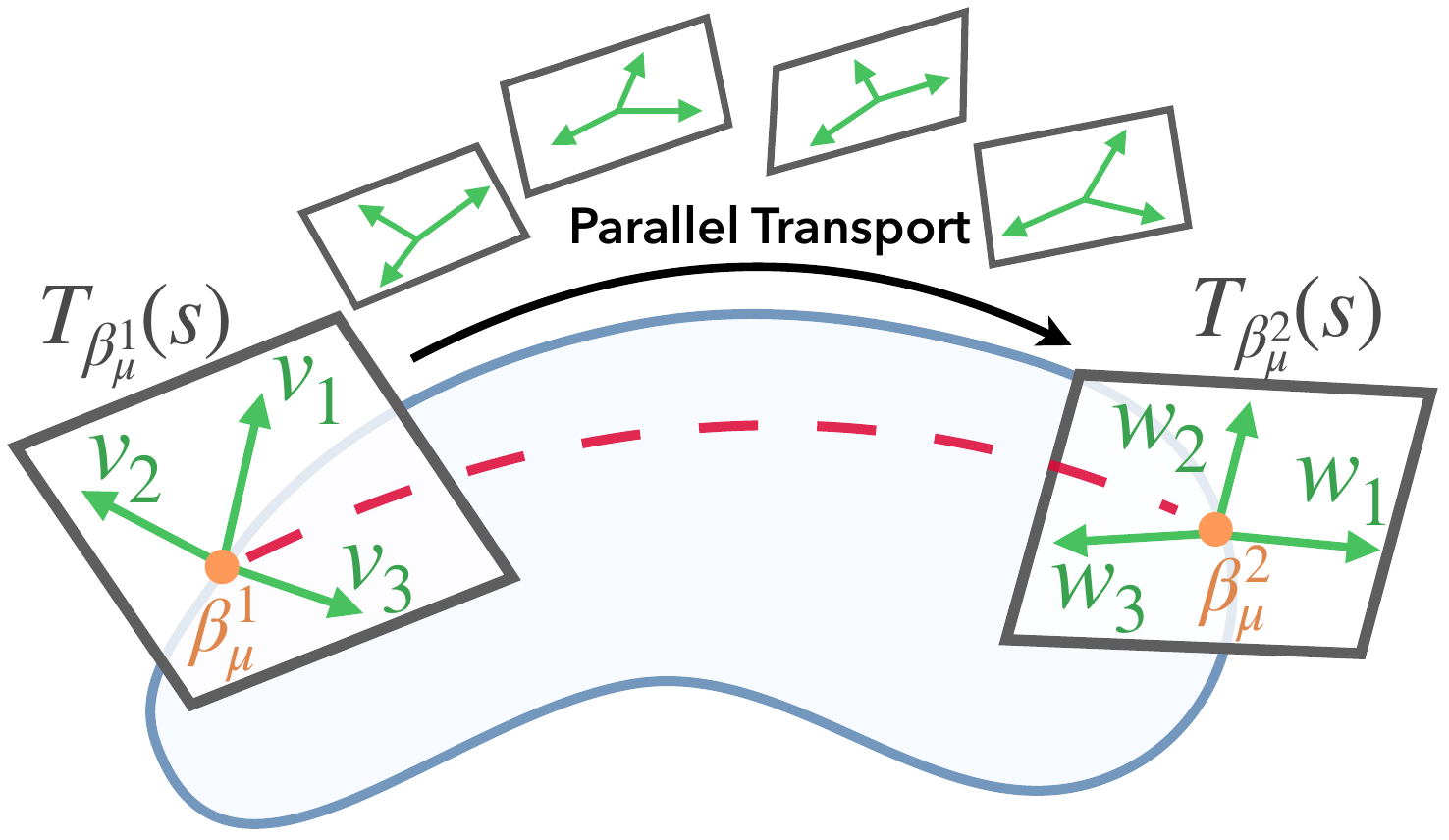}

\caption{A visualization of the parallel transport process  of one tangent space to another.} 
\label{fig:schematic}

\end{figure}

\begin{algorithm}[tbh]
\caption{Parallel Transport of $\{v^2_i\}$ for the reconstruction of $\aproj^2$ along a geodesic from $\beta_\mu^2$ to $\beta_\mu^1$}
\label{algo:parallel_tx}
 \SetAlgoLined
  \KwIn{$(\beta_\mu^1, \{v^1_i\}, i = 1, \ldots, N)$, $(\beta_\mu^2, \{v^1_i \}, i = 1, \ldots, N)$} 
  \KwOut{Transported tangent vectors $\{\tilde{v}^2_i\}, i = 1, \ldots, N)$}
  
    Compute a tangent vector $w$ such that  $\mbox{exp}_{\beta_\mu^2}(w) = \beta_\mu^1$
    
  	Let $l_w = \sqrt{\langle w,w \rangle}$
  	
  	Define a step size $k$.
  
    \For{$\tau \gets2$ \KwTo $k-1$ }{
		$q_{\tau} = \mbox{exp}_{\beta_\mu}(\frac{w}{k})$
	
		$\tilde{v_i} = v_i - \langle v_i, q_{\tau} \rangle
		q_{\tau}$, $i = 1, \ldots, N$
		
		$\tilde{v_i} = \tilde{v_i}
		\frac{l_w}{\norm{\tilde{v_i}}}$, $i = 1, \ldots, N$

    }
    
 \end{algorithm}

%%%%%%%%%%%%%%%%%%%%%%%%%%%%%%%%%%%%%%%%%%%%%%%%%%%%%%%%%%%%%%%%%%%%%%%%%%%%%%%%%
We can formalize our approach by stating it as the following metric between two tracts $\bundl^i = (\beta_\mu^i, \aproj^i)$,~ $\bundl^j = (\beta_\mu^j, \aproj^j)$ by equipping the induced product-space metric:

\begin{equation}
\label{eq:dist_func}
D(\bundl^i, \bundl^j) = \inf_{\gamma, \mathcal{O}} \sqrt{{||\beta_{\mu}^i - \sqrt{\dot{\gamma}}\beta_{\mu}^j(\gamma)||}^2 + {||\aproj^j - \mathcal{O}(\aproj^i)||_{fro}}^2}
\end{equation}

 With some abuse of notation, $\aproj^i$, is the low dimensional tangent vector representation after parallel transport. The metric on the low-dimensional features is given by the Frobenius metric. By determining the optimal rotation $\mathcal{O} \in SO(N)$ and diffeomorphism $\gamma:[0,1] \rightarrow [0,1]$, we get i) a registration between the two tract streamlines and ii) a measure of similarity between tract streamlines. Since some shape features may be lost in the low-dimensional setting, we refer to this alignment as a ``soft" deformation. Note that a ``hard"-alignment can be performed if one then takes the assigned tangent vectors after the soft deformation and applies the warping process to nearest neighbors of tangent vectors.

%%%%%%%%%%%%%%%%%%%%%%%%%%%%%%%%%%%%%%%%%%%%%%%%%%%%%%%%%%%%%%%%%
\section{Results}
\subsection{Data}

\subsubsection{Image Acquisition and Preprocessing}
We obtained diffusion-weighted and structural images using a 3T Siemens PRISMA scanner and a $32$-channel head coil from $n=43$ healthy adult subjects (Age: 32.2 $\pm$ 11.8, Sex: 21M/22F). Diffusion-weighted images were acquired using a spin-echo echo planar sequence (EPI), which included 14 reference images ($b=0$ s/mm$^2$), and multishell images ($b=1500, 3000$ s/mm$^2$) with 92 gradient directions along with a T1-weighted image (voxel size=0.8mm$^3$). We followed the Human Connectome Project minimal preprocessing pipeline to process the imaging data~\cite{glasser2013minimal}. T1-weighted structural images were registered to Montreal Neurological Institute (MNI) 152 T1 standard space for anterior commissure - posterior commissure (AC-PC) alignment. 12-degree of freedom (DOF) registration of individual T1 structural image to MNI T1 template was first performed, followed by the 6-DOF registration of individual image to the 12-DOF individual-to-MNI-template registered image. Then, each subject's diffusion weighted image was registered to the structural image using 6-DOF registration~\cite{glasser2013minimal}. 

\subsubsection{Tractography and Along-tract Diffusion Measure Extraction}
Whole brain tractography was performed with MRtrix3~\cite{TOURNIER2019116137}, using multi-shell multi-tissue constrained spherical deconvolution followed by filtering of tractograms~\cite{smith2015sift2} to obtain more biologically relevant fibers, ultimately producing 10 million fibers for each subject~\cite{JEURISSEN2014411}.  Whole brain fiber tracts were segmented into 18 fiber groups; Left (L) and Right (R) Thalamic Radiation (Th Rad), Corticospinal Tract (CST), Cingulate Cingulum (CnCn), Cingulate Hippocampus (CnHp), Superior/Inferior Longitudinal Fasciulus (SLF/ILF), Uncinate (Unc), and Arcuate (Arc), and Corpus Callosum Forceps Major/Minor (CC F Maj/Min) using Automated Fiber Quantification (AFQ)~\cite{Yeatman2012}. We selected an arbitrary subject as a representative and performed bundle-wise alignment of all subjects to this template. 

\subsubsection{Execution time}
Computations of the experiments were performed on a 2.6 GHz 6-Core Intel Core i7 MacBook Pro. The tract-to-tract registration averaged to 33.77 minutes per registration, with K{\"a}rcher mean contributing to approximately 18.48 minutes of the computational time.

\subsection{Within-tract Fiber Alignment to the K{\"a}rcher mean}
\begin{figure}[htb]
\centering
    \includegraphics[width=0.87\textwidth]{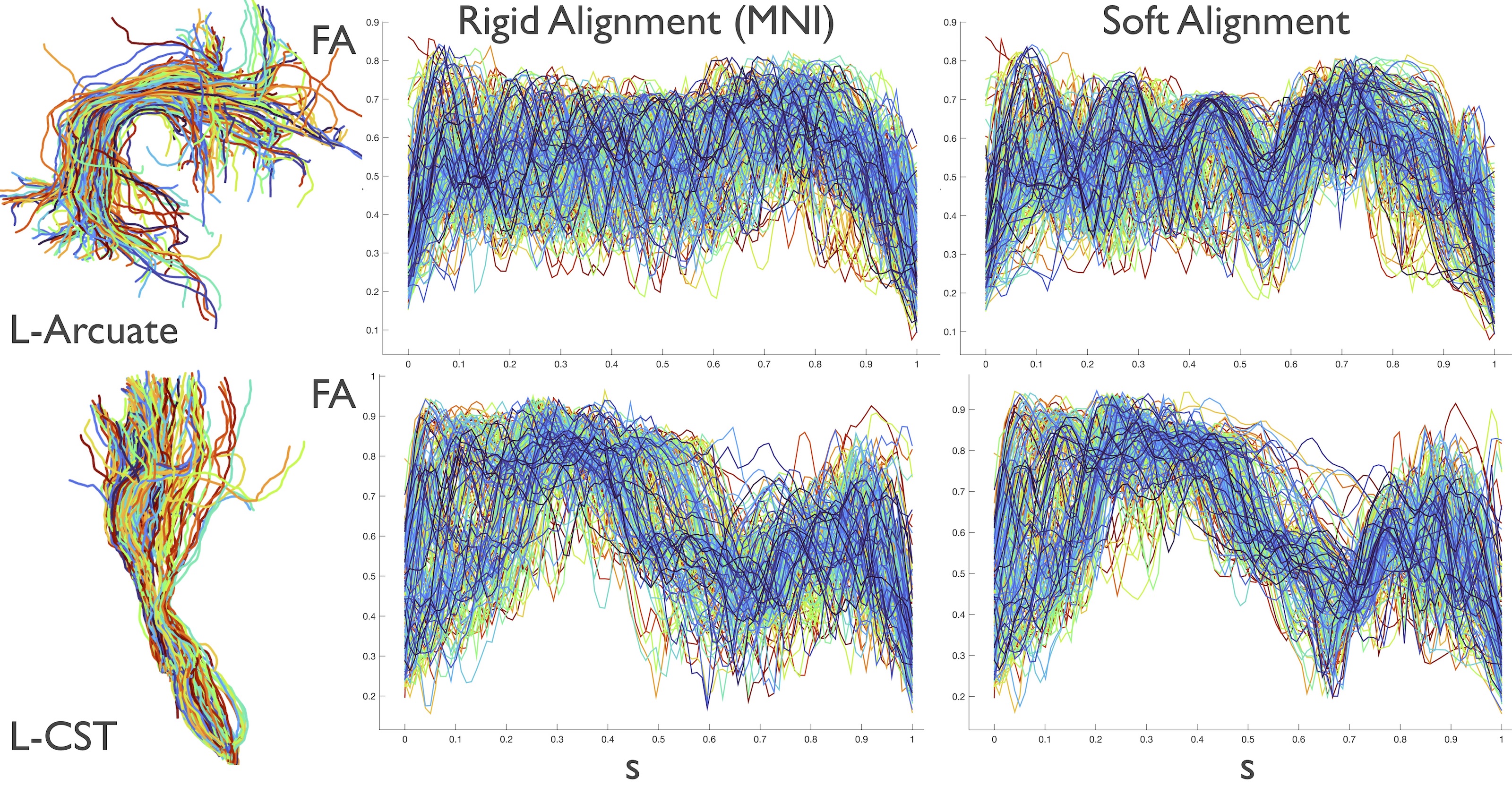}
 \caption{Left: Fiber tracts of the L-Arcuate and the L-CST. Center: Along-tract FA profiles for all fibers  in the respective bundles obtained from the initial (rigid) alignment. Right:  Re-parameterized FA profiles after aligning each fiber to the within-tract K{\"a}rcher mean.
} \label{fig:FA_within_bundle}

\end{figure}

Prior to performing subject-to-subject tract alignment, we demonstrate the effectiveness of the within-bundle fiber alignment to the mean by observing the profiles of along-tract fractional anisotropy for all fibers in that tract. Figure \ref{fig:FA_within_bundle} shows FA profiles after the initial rigid alignment and after undergoing reparameterization resulting from the soft alignment of the L-Arcuate and the L-CST tracts to the template. It is observed that the FA profiles show distinct grouping of features and visually shows reduced variability along the length of the tract, even though the objective function for the soft alignment process did not explicitly incorporate FA as a measure. FA profile alignment is typically used as a qualitative descriptor indicating that a tract is properly aligning to the mean, and we expect to see noisy profiles to form more regular patterns if a registration process is properly aligning the tract. Figure \ref{fig:FA_within_bundle} demonstrates how the process of deformation transfer via parallel transport and rotation corresponds to smoother and more regular FA-profiles suggesting that this technique is in fact improving registration over rigid alignment. 

\subsection{Soft-alignment of Fiber-tracts}
\begin{figure}[tbh]

\centering
     \includegraphics[width=0.87\textwidth]{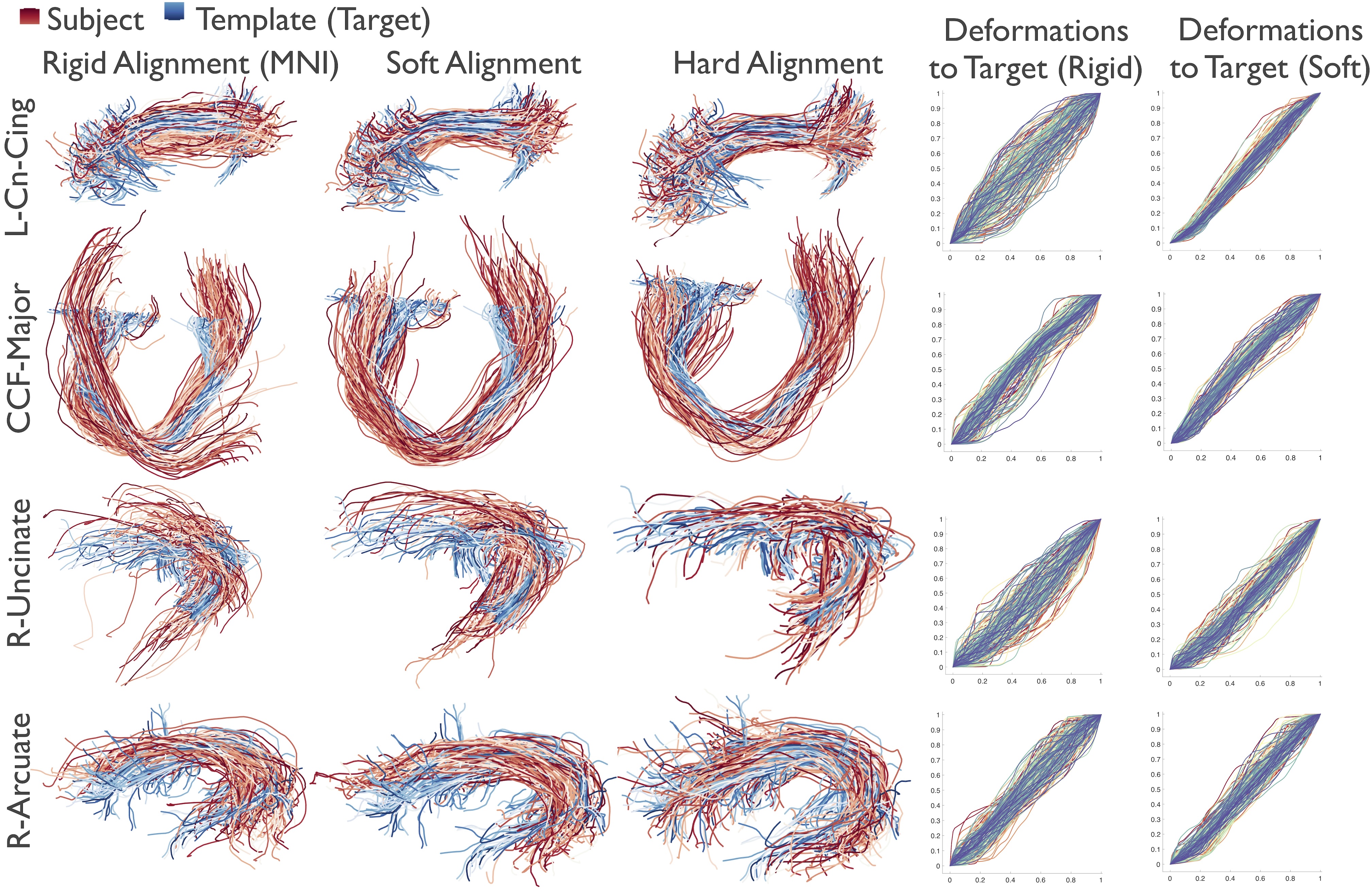}

\caption{Comparison of soft and hard alignment between a subject and a template along with warping functions ($\gamma$) against rigid alignment in the MNI space. } 
\label{fig:align_bundles}

\end{figure}

 Figure~\ref{fig:align_bundles} shows examples of rigid  alignment, soft alignment, and hard alignment of L-Cn-Cing, CCF-Major, and R-Uncinate bundles between a subject and the template. The last two columns show the deformation represented by the warping functions $\gamma$ between the individual fibers of the subject, and the target fibers of the template. Note that the closer these $\gamma$ functions are to identity function, corresponds to individual fibers in the subject tract being more aligned to the template.
 
 Although the initial rigid alignment brings the subject closer to the template, one can observe discrepancies in the overall geometry of the fibers. Additionally, the warping functions deviate further from the identity indicating that a majority of the fibers are not aligned. The soft alignment improves upon the overlap by achieving an overall agreement of the shapes. We visually see that fibers from the subject tract conform to the overall template shape, and that the warping functions have aligned closer to the identity.
 
 The hard alignment improves upon the soft-alignment by further refining over the local details and especially respecting the ends of the fiber tracts. In the case of CCF-Major (middle-row), hard-alignment succeeds in warping the subject fibers to conform to the overall shape of the template. Note, while one can directly pairwise register the fibers between subject and template, the initial pairing of the subject fibers and the template fibers is arbitrary. Thus the warping between such fibers incurs a large cost of reparameterization. 
 
\subsection{Tract Point-set Similarity}
There is no direct method for comparing the closeness of tract shapes between the subject and the target, as the rigid-alignment does not provide comparable fiber correspondences between tracts unlike the soft alignment method. To overcome this challenge, we compute the bidirectional Hausdorff distance that can be applied to arbitrary point-sets~\cite{huttenlocher1993comparing}. This is given by $D_H(\bundl_1, \bundl_2) = \max (d_H(\bundl_1, \bundl_2), d_H(\bundl_2, \bundl_1)) $, where $d_H(\bundl_1, \bundl_2) = \max_{a \in \bundl_1} \min_{b \in \bundl_2} || a - b||$.  
The bidirectional Hausdorff distance $D_H$ was computed for each tract for $N=43$  subjects and was shown to be  significantly less after soft alignment for all tracts ($p < 1e-5$) except for the CC F Min ($p=0.0748$) after correcting for multiple comparisons using FDR~\cite{benjamini1995controlling}. The average Hausdorff distance for CC F Min was still lower for the soft alignment method but did not survive FDR.

\begin{figure}[tbh]

\centering
     \includegraphics[width=1.0\textwidth]{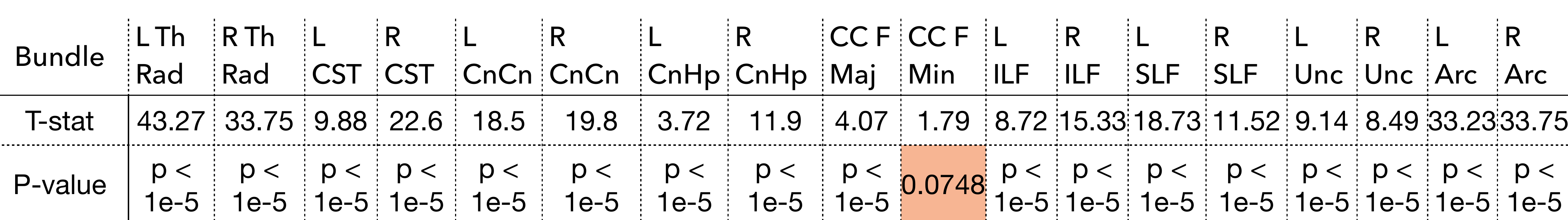}

\caption{Comparisons of tract point-set closeness between rigid and soft alignment (significantly smaller) using Hausdorff distance after FDR correction for $N=43$ subjects. The CC F Min tract (shaded) did not achieve significance. } 
\label{fig:Hausdorff}
\end{figure}

\section{Discussion}
We presented a novel framework for soft registration of white matter fiber tracts using a low-dimensional representation that encodes shape deformations. The mechanism of parallel transport and product metric enables an effective computation of tract differences while simultaneously allowing the alignment of tracts. From within-tract fiber-to-mean registration results, we see that the shape alignment of geometrically similar fibers may enhance the features of diffusion measures sampled along their lengths even though the measure (FA) was not explicitly accounted for in the deformation process. This framework is general and will potentially allow statistical shape analysis of general collections of streamlines.

\section{Acknowledgements}
This research was partially supported by a fellowship from the NSF NRT Award \#$1829071$ (EN) and the NIH NIAAA (National Institute on Alcohol Abuse and Alcoholism) awards R01-$AA025653$ and R01-$AA026834$ (SHJ). Data acquisition and processing was also supported by NIH/NIMH award U01MH110008.

%%%%%%%%%%%%%%%%%%%%%%%%%%%%%%%%%%%%%%%%%%%%%%%%%%%%%%%%%%%%%%%%%%%%%%%%

%For citations of references, we prefer the use of square brackets
%and consecutive numbers. Citations using labels or the author/year
%convention are also acceptable. The following bibliography provides
%a sample reference list with entries for journal
%articles~\cite{ref_article1}, an LNCS chapter~\cite{ref_lncs1}, a
%book~\cite{ref_book1}, proceedings without editors~\cite{ref_proc1},
%and a homepage~\cite{ref_url1}. Multiple citations are grouped
%\cite{ref_article1,ref_lncs1,ref_book1},
%\cite{ref_article1,ref_book1,ref_proc1,ref_url1}.
\bibliographystyle{splncs04}
\bibliography{paper}
\end{document}